%% file: main.tex
\PassOptionsToPackage{dvipsnames,table}{xcolor}
\documentclass[10pt,twocolumn,letterpaper]{article}

\usepackage[pagenumbers]{cvpr} 
\usepackage[table]{xcolor}
\usepackage{multirow} 
\usepackage{multicol}
\input{preamble}

\definecolor{cvprblue}{rgb}{0.21,0.49,0.74}
\usepackage[pagebackref,breaklinks,colorlinks,allcolors=cvprblue]{hyperref}

\def\paperID{61} 
\def\confName{CVPR}
\def\confYear{2025}

\title{Improving Physical Object State Representation in Text-to-Image Generative Systems}
\author{%
  Tianle Chen\textsuperscript{1}\quad
  Chaitanya Chakka\textsuperscript{1}\quad
  Deepti Ghadiyaram\textsuperscript{1,2}\\[0.5em]
  \textsuperscript{1}Boston University\quad
  \textsuperscript{2}Runway\\[0.5em]
  \{{tianle, chvskch, dghadiya}\}@bu.edu%
}
\begin{document}
\input{sec/figs/fig1}
\input{sec/0_abstract}    
\input{sec/1_intro}
\input{sec/2_relatedwork}
\input{sec/figs/fig2}
\input{sec/3_approach}
\input{sec/4_experiments}
\input{sec/5_conclusion_future_work}
{
    \small
    \bibliographystyle{ieeenat_fullname}
    \bibliography{main}
}
\input{sec/appendices}

\end{document}

%% file: preamble.tex
\usepackage[normalem]{ulem}

%% file: sec/figs/fig1.tex
\twocolumn[{
\maketitle
\begin{center}
\newcommand{\teaserwidth}{0.82\textwidth}
    \includegraphics[width=\teaserwidth]{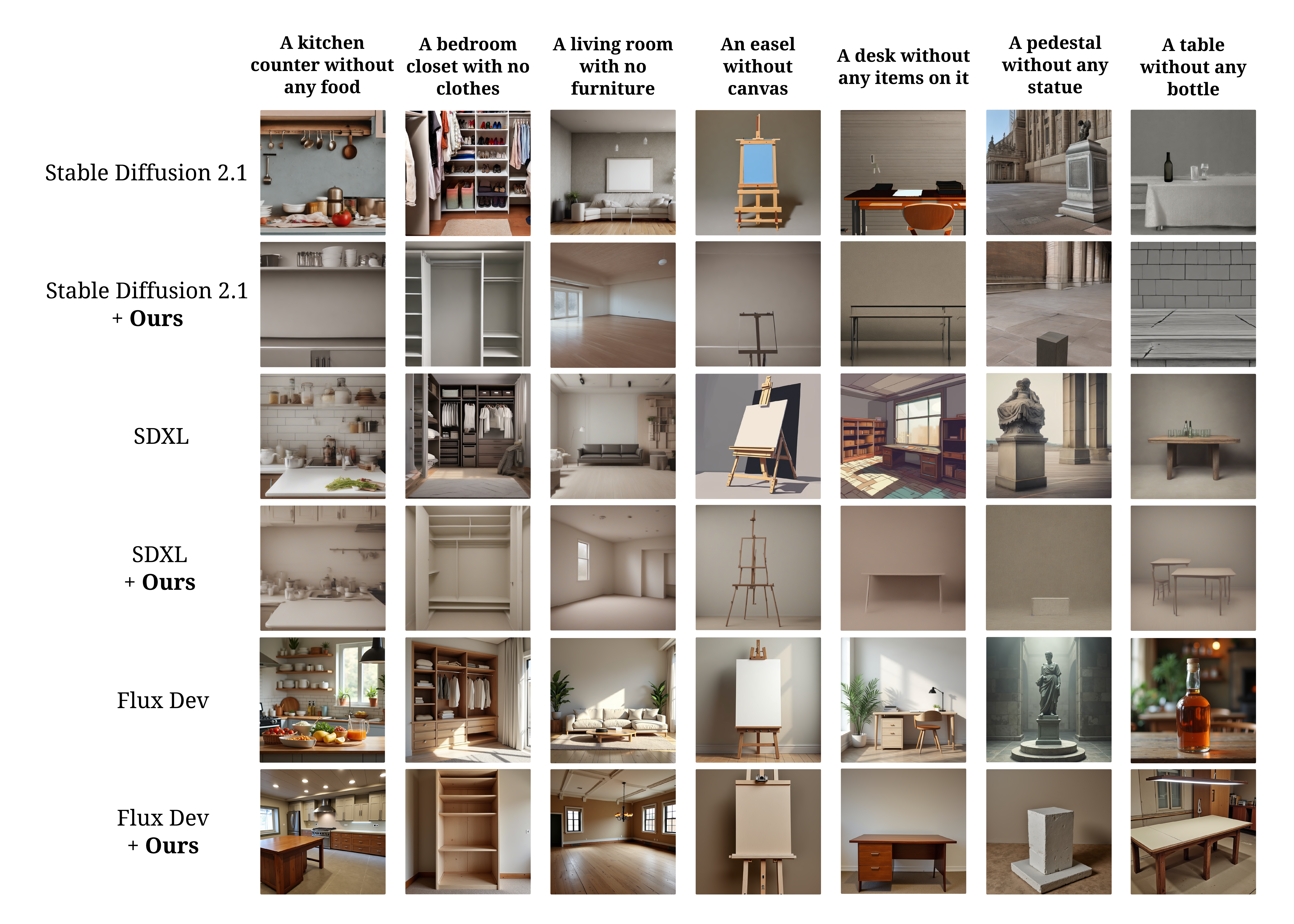}
\vspace{-2.2mm}
\captionof{figure}{{\textbf{Current text-to-image models struggle to depict common objects in varied physical states}, inaccurately include unintended objects or fail to depict the requested empty or absence state (e.g., prompting for ``A kitchen counter without any food" still results in a kitchen count full of food). Our method addresses these issues and yields accurate object state representation. }}
\label{fig:hero_figure}
\end{center}
}]

%% file: sec/0_abstract.tex
\begin{abstract}
Current text-to-image generative models struggle to accurately represent object states (e.g., ``a table without a bottle,'' ``an empty tumbler''). In this work, we first design a fully-automatic pipeline to generate high-quality synthetic data that accurately captures objects in varied states. Next, we fine-tune several open-source text-to-image models on this synthetic data. We evaluate the performance of the fine-tuned models by quantifying the alignment of the generated images to their prompts using GPT4o-mini, and achieve an average absolute improvement of \textbf{8+\%} across four models 
 on the public GenAI-Bench dataset. We also curate a collection of 200 prompts with a specific focus on common objects in various physical states. We demonstrate a significant improvement of an average of \textbf{24+\%} over the baseline on this dataset. We  release all evaluation prompts and code at \url{https://github.com/cskyl/Object-State-Bench}.
\end{abstract}

%% file: sec/1_intro.tex
\section{Introduction}
\input{sec/figs/advanced_model_failure}
Recent advances in text-to-image generation~\cite{Rombach_2022_CVPR, diffusion1,diffusion2, flux2024, dalle3, xiao2024omnigen, team2023gemini, sora, runway, wang2024emu3, chen2025janus} have significantly improved the visual quality and correctness of the generated images and unlocked incredible potential for human creative expression. Despite these significant improvements, as illustrated in Fig.~\ref{fig:advanced_fail}, existing generative and visual language models still face challenges in accurately capturing spatial relationships~\cite{spatial} and simple real-world physical states such as presence, absence, empty, full~\cite{sun_object}, and so on. Despite being trained on billions of images, recent studies~\cite{zhao2024lost} and examples in Fig.~\ref{fig:advanced_fail} suggest that the generative systems are latching onto the intended and unintended co-occurring contexts represented in the training data and lack a fundamental understanding of object states. 

Let us consider the example of a kitchen shelf. A simple web search indicates that a significant portion of images of a kitchen shelf are typically shelves filled with a variety of objects. We posit that this may inadvertently induce \textit{contextual bias} into both generative and vision language models trained on such data, leading to text-to-image systems mostly generating a kitchen shelf in a ``full'' or ``occupied'' state. This scenario is further exacerbated given that captions generated on training data usually capture objects \textit{present} in the image and not the ones \textit{absent} in it. Recent studies~\cite{CLIP_negation, VLM_Negation} have shown evidence that CLIP~\cite{OpenAI_CLIP}, the defacto text encoder in most generative image models, struggles to understand negation. How then should we impart the concept of absence of an object (e.g., a table without a vase) or negation of a physical state (e.g., an empty bottle) to the text to image generative system?

We tackle this issue in our work. Our key idea is to supply a text to image model with more evidence of a variety of objects in diverse physical states during training so that the model implicitly learns what absence or negation of an object should visually look like. We do this by first designing an automated pipeline to generate high-quality synthetic data that explicitly captures daily objects in varied, naturally feasible object states. The data generation pipeline (Fig.~\ref{fig:fig2_pipeline}) comprises a step involving generating template-like prompts describing common objects in different physical states. Next, we use an off-the-shelf text to image model to generate synthetic images and subsequently filter out those not representative of objects in absent or empty states using a vision language model. We then finetune publicly available text to image models on this diverse generated data.

Our experiments indicate that finetuning on this curated synthetic dataset enriches a model's understanding of objects and their various physical states. Furthermore, we probe if finetuning leads to a more holistic understanding of objects in diverse physical states or mere memorization. To this end, we test the model on prompts comprising objects not part of the finetuning data. Across the four models we evaluate, we observe an average improvement \textbf{24.6\%} on novel, unseen objects, indicating that finetuning led to a better, more generalizable latent space. Qualitative comparisons (Fig.~\ref{fig:hero_figure}) further illustrate these improvements. Beyond enhancing the overall capability of generating objects in diverse states, we also examine the impact of fine-tuning on visual quality. Our analysis shows that both the CLIP score and the FID remain at similar levels after fine-tuning, with the FID showing no significant change, confirming that our synthetic data does not introduce undesirable visual artifacts. We also conduct a user study that further confirms that visual quality remains perceptually indistinguishable (see Appendix~\ref{sec:appendix_user_study}). We also show that fine-tuning on high-quality synthetic data does not degrade performance on other prompts not related to object states. 
We summarize our key contributions below:
\begin{itemize}[noitemsep,topsep=0pt]
\item We propose a \textbf{fully-automatic synthetic data generation pipeline} to systematically create high-quality training data that explicitly targets objects in empty, negation, and absent states (Sec.\ref{sec:approach}).
\item We fine-tune four open-source text-to-image generative models on this synthetic data and evaluate the generations on the publicly available GenAI-Bench dataset ~\cite{li2024genai}. We show that the finetuned models yield an improvement of averaging \textbf{8+\%} across the four models we experimented with, as measured by quantifying the alignment of the generated images to their prompts (Sec.\ref{sec:experiment}).
\item We introduce a novel prompt collection, \textbf{Object State Bench}, specifically designed to evaluate models in complex physical states. All finetuned models yield an overall improvement of an average of \textbf{24+\%} over their untuned baselines on this benchmark. This strong result underscores the substantial potential of our synthetic data pipeline in addressing the critical limitation of capturing object states in text-to-image generation.  Both resources are publicly available at \url{https://github.com/cskyl/Object-State-Bench}.
\end{itemize}

%% file: sec/figs/advanced_model_failure.tex
\begin{figure*}[t]
\centering
\includegraphics[width=0.8\linewidth]{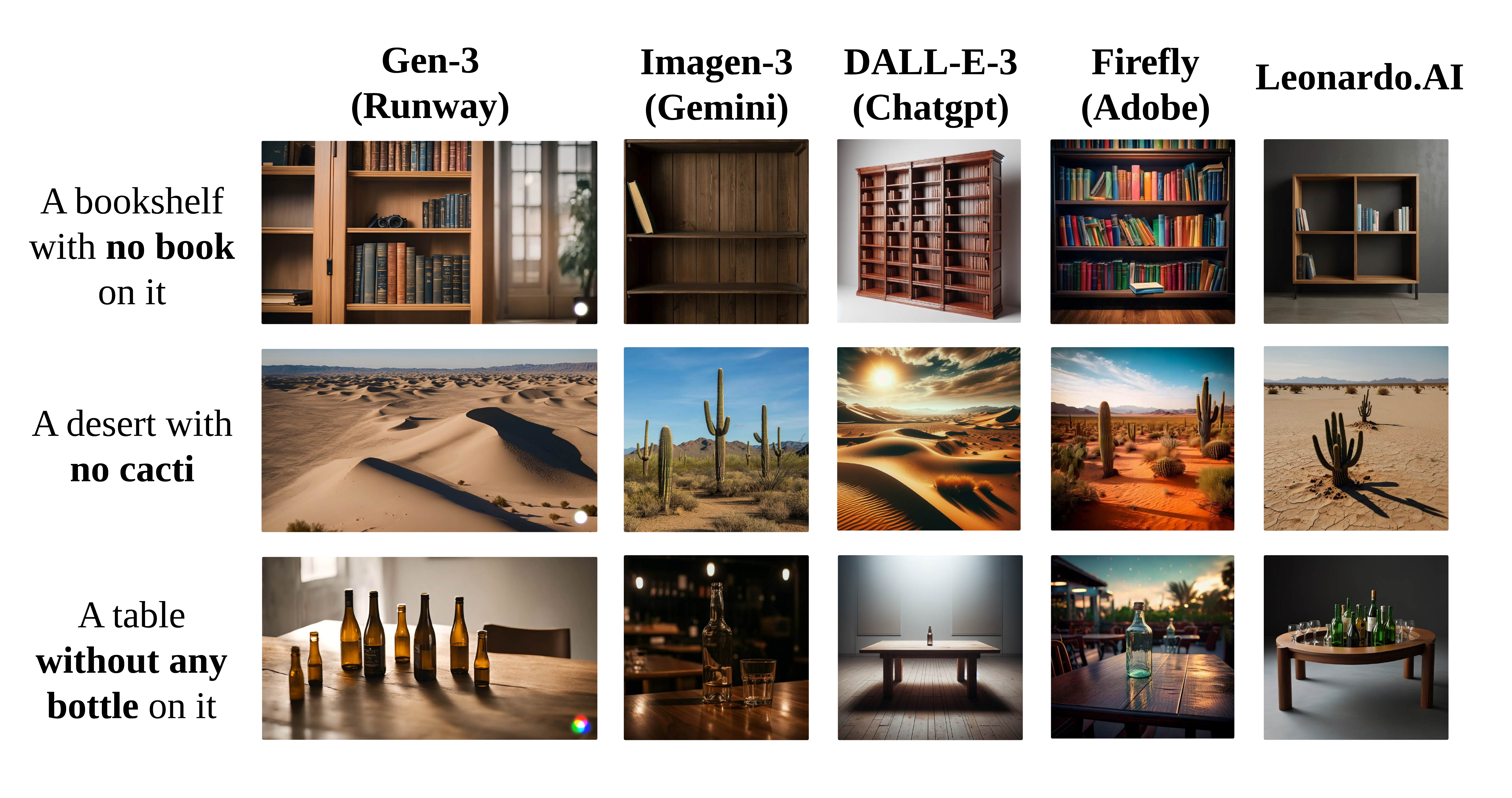}
\vspace{-0.2in}
\caption{\textbf{State-of-the-art closed-source text-to-image and text-to-video models struggle to depict objects in absent or negation states.} For Gen‑3 (a text-to-video model), we show a single extracted frame. This highlights the limitations of current advanced generative systems in accurately representing objects in simple and common physical states.}
\label{fig:advanced_fail}
\end{figure*}

%% file: sec/2_relatedwork.tex
\section{Related Work}
\noindent \textbf{Iterative Correction and Guidance Techniques:}  
Recent studies have addressed shortcomings in text-to-image synthesis by incorporating iterative correction mechanisms. For example,~\citet{wu2024self} propose a Self-correcting LLM-controlled Diffusion framework that leverages large language models to iteratively refine generated images through latent space operations such as addition, deletion, and repositioning. Similarly,~\citet{liu2024correcting} introduces a particle filtering framework that uses external guidance, such as object detectors and real images, to mitigate errors such as missing objects and image distortions. Although these methods improve image fidelity, they typically require additional inference-time computations and complex feedback loops.
\noindent \textbf{RL-based Preference Optimization Methods:} 
In parallel, reinforcement learning (RL) techniques have been explored to fine-tune generative models for better alignment with human preferences. Methods such as Direct Preference Optimization (DPO)~\cite{wallace2024diffusiondpo} and Proximal Policy Optimization (PPO)~\cite{ren2024diffusion} have been applied to adjust model outputs based on preference feedback. These approaches optimize the generation process by modifying latent representations or output distributions to prioritize images that not only exhibit high visual fidelity but also maintain semantic accuracy. Although initially developed for language models, recent adaptations of these RL-based methods to stable diffusion models have shown promising improvements. However, such techniques often involve complex training procedures and significant computational overhead. Our work complements these efforts by adopting a data-centric approach that focuses on synthetic data generation.

\noindent \textbf{Synthetic Data Generation:}  
Complementary to iterative correction and RL-based tuning methods, synthetic data generation has emerged as a promising strategy to overcome limitations in training datasets. Recent work~\cite{lomurno2024stable} demonstrates that synthetic data produced via diffusion models can enhance model robustness and generalization. Our method builds on this idea by adding a fully automatic pipeline that creates high-quality synthetic image-prompt pairs. These pairs clearly show different object states that aren't always shown in real-world data. Distinct from methods relying on iterative correction or resampling during inference, our pipeline directly enhances the training process. We fine-tune open-source text-to-image models on our synthetically generated data, thereby improving the semantic alignment between generated images and their textual descriptions. This data-centric framework not only simplifies the overall generation pipeline but also offers an extensible solution to address critical semantic limitations in text-to-image synthesis.

%% file: sec/figs/fig2.tex
\begin{figure}[t!]
    \centering
    \includegraphics[width=1.05\linewidth]{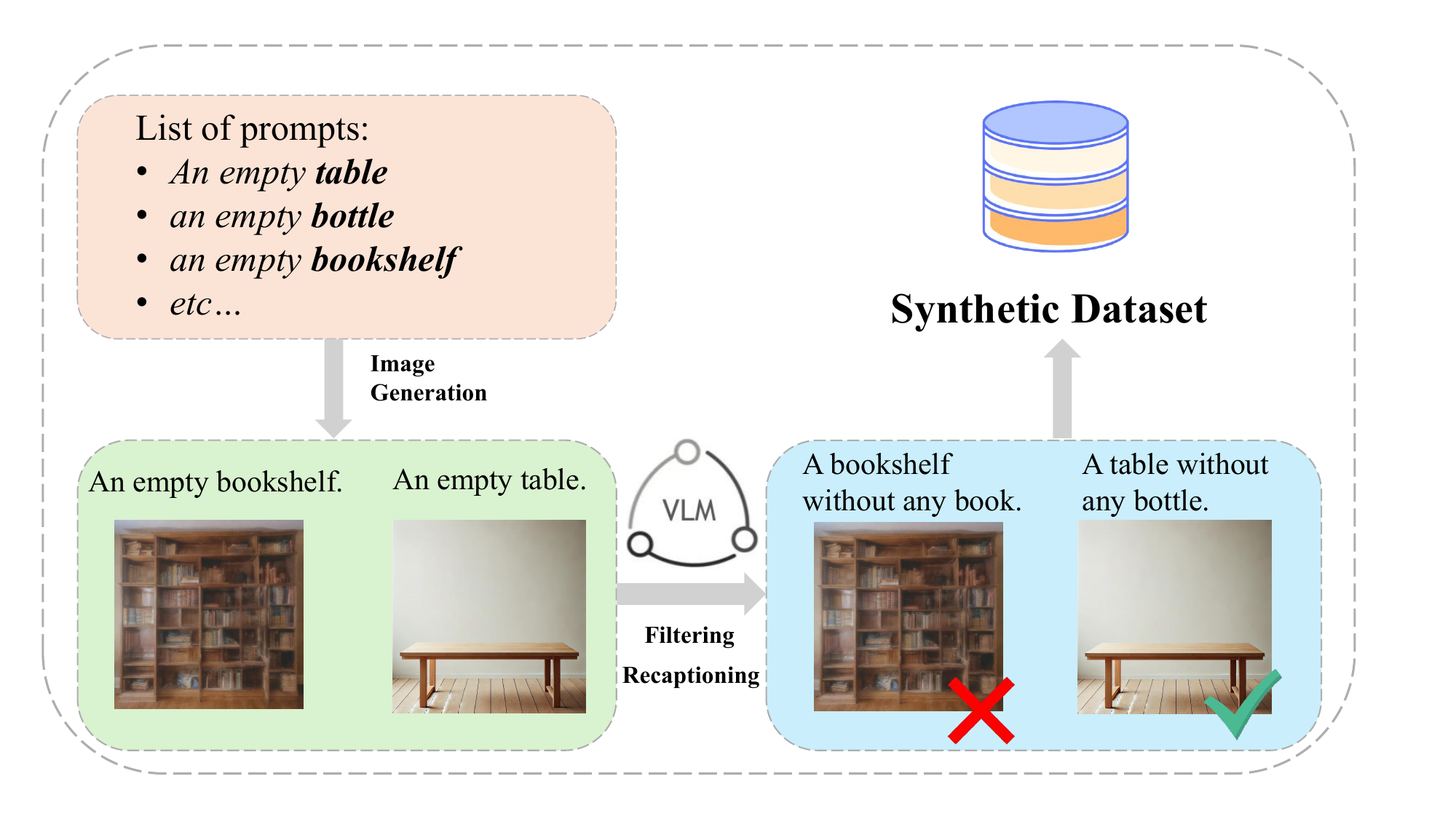}
    \caption{
    \textbf {Overview of the proposed synthetic data generation pipeline:} We generate prompts describing common objects in different physical states. We next create images from the prompts, evaluate for the correct representation of the object state using GPT4o-mini~\cite{hurst2024gpt}. We rephrase prompts to introduce diversity in the sentence structures, length, and objects specified.}
    
    \label{fig:fig2_pipeline}
    \vspace{-1em} 
\end{figure}

%% file: sec/3_approach.tex
\section{Approach} \label{sec:approach}
To address the gap in accurately generating objects in common physical states using text-to-image models, we propose a fully automatic synthetic data generation pipeline. As illustrated in Fig.~\ref{fig:fig2_pipeline}, we first identify a diverse set of real-world objects and compose prompts referring to those objects in different physical states. Next, we generate images corresponding to those prompts using an off-the-shelf text to image generative model. Following this, we leverage Large Language Models (LLMs) to refine prompts and introduce diversity in their syntax. We also use a Large Vision-Language Model (LVLMs) for visual verification of the object state representation, to reduce noise in the generated data. We describe each step in detail next.

\noindent \textbf{Noun identification, prompt, and image generation:} First, we use a large language model and curate a wide range of real-world objects such as containers, tables, shelves, rooms, and drawers, that can be depicted in empty, full, and absent states. 
We next use a large language model to compose simple prompts around these objects that explicitly describe empty states (e.g., ``an empty table"). We note that in our work, we focus only on curating prompts and images that represent only empty object states as they are more commonly underrepresented in most datasets.

\input{sec/figs/filtering_prompt}
\noindent \textbf{Image Synthesis and Filtering:}
Next, we utilize an off-the-shelf text-to-image generation model to produce multiple candidate images per prompt using random seeds to ensure sufficient content diversity. Given that existing models 
struggle to generate images that represent object states correctly, we pass each generation through a vision language model, and use visual-question-answering prompts to filter out generations that do not capture the object states correctly

\noindent \textbf{Recaptioning:} Finally, we use a large language model to introduce more diversity into the template-like initial prompt syntax. For example, the initial prompt ``an empty table" could be refined to ``a table without any bottle or book," increasing the complexity and clarity of the generated prompts. Our full pipeline is depicted in Fig.~\ref{fig:fig2_pipeline}. 

We finetune several open-sourced models on this synthetically generated image data, which we describe next.

%% file: sec/figs/filtering_prompt.tex
\begin{figure}[t!]
    \centering
    \fbox{
        \begin{minipage}{0.95\linewidth}
            \textbf{Prompt for filtering generated images not representing absence or empty state of an object:}
            \vspace{0.3cm}
            \textit{\\``You are an assistant that evaluates whether an image correctly represents the `empty state' of an object as described in the caption. Specifically, check if the main object appears empty or unoccupied and confirm that the described absent object is not present in the image. Does the image accurately reflect both conditions? Return `Yes' or `No'."}
        \end{minipage}
    }
    \caption{\textbf{System prompt used} on the generated images for filtering out images not aligning with the provided prompt.} 
    \label{fig:filtering_prompt}
\end{figure}

%% file: sec/4_experiments.tex
\section{Experiments} \label{sec:experiment}
This section briefs about the datasets and evaluation metrics, baseline models, followed by fine-tuning. We also study the effect of different design choices we make in our overall approach.
\input{sec/figs/sd15_comparision}
\input{sec/figs/sd21_comparision}
\input{sec/figs/sdxl_comparision}
\input{sec/figs/flux_comparision_fig}
\input{sec/figs/omnigen_comparision_fig}
\input{sec/4_1_implementation_details}

\input{sec/4_2_fine_tuning}

\subsection{Overall performance improvements}
Table~\ref{tab:baseline_performances} presents GPT and VQA results for five open-source text-to-image models before and after fine tuning with our synthetic dataset. Fine tuning raises GPT scores in average by 8.2 percentage points on GenAI Object State and by 24.6 points on Object State Bench, while VQA accuracy increases by 5.2 and 17.2 points, respectively. The most substantial gains occur for the SDXL model, which reaches 23\% GPT score and 58\% VQA score on the Object State Bench. These consistent improvements demonstrate that our synthetic data pipeline and fine-tuning approach effectively enhance semantic alignment and visual question answering performance across diverse models.
Furthermore, qualitative comparisons ( Fig.~\ref{fig:SD15_compare}, \ref{fig:sd21_compare}, \ref{fig:sdxl_compare}, \ref{fig:flux_compare} and \ref{fig:omnigen_compare}) visually illustrate the enhanced semantic alignment in the generated images after fine-tuning on samples from both the GenAI-object-state and Object State Bench datasets.

While our fine-tuning strategy generally improves the visual representation of object states, qualitative analysis reveals several different failure modes, as illustrated in Fig.~\ref{fig:qualitative_tuning}.  In Fig.~\ref{fig:improvement_case}, the tuned model improves significantly by approaching the correct empty state. However, the representation remains imperfect, indicating that the model converges toward the intended state without fully capturing all semantic details.  In contrast, Figure~\ref{fig:overrepresentation_case} shows instances where the tuning process overemphasizes emptiness, resulting in an overrepresentation of the empty state, which, causes the object itself to be imperfectly represented.  These observations highlight the delicate balance required in fine-tuning: while reinforcing the concept of emptiness is beneficial, overemphasis can degrade the precise representation of the object.  
\subsection{Effect of the synthetic data generator}
Table~\ref{tab:results_synthetic} presents GPT evaluation scores for three generative models fine tuning on synthetic data produced by different generators. Overall, the choice of generator has only a modest effect. For Stable Diffusion 1.5, using SDXL data yields the highest GenAI-Object-State score at 24\%, while Stable Diffusion 2.1 data delivers the top Object State Bench result of 56\%. In the case of SDXL, both SDXL and Stable Diffusion 2.1 generators achieve the best bench score (58\%), with Stable Diffusion 1.5 data slightly outperforming them on GenAI-Object-State (23\%). Similarly, tuning Stable Diffusion 2.1 with Stable Diffusion 1.5 or SDXL data produces identical GenAI-Object-State scores of 23\% and bench scores between 55\% and 56\%, while its own data trails by one point on both benchmarks. These small variations demonstrate that our synthetic data pipeline is robust: regardless of the underlying generator, all tuned models show consistent improvements in semantic alignment on both evaluation sets.  

\input{sec/tables/finetune_performances}
\input{sec/figs/partial_improvement_case}
\subsection{Generalization to unseen object}
Our overarching goal is to teach the model the concept of emptiness or the absence of an object. To this end, we study whether the model’s understanding of object states learned during training generalizes to novel, unseen objects. We first identify $100$ novel objects that are not part of the $3000$ objects used for training. Furthermore, we manually inspect the list of objects and filter out if the new object is similar to the training objects (see Appendix~\ref{appendix:object_list} for the complete list of objects). Using our data generation pipeline described in Sec.~\ref{sec:approach}, we generated images for these objects.

The results, reported in Table~\ref{tab:generalizability_results} (Column 2), show an improvement in the GPT score from 13.0 to 20.0 (+7 percentage points).
Such significant improvements on unseen objects indicates that finetuning on synthetic data of objects in absent and negation states is leading to a more comprehensive understanding of physical states even on novel, unseen objects.
\input{sec/tables/generalizability_performances}


\subsection{Performance on non-empty object states}
Given our finetuning data consists of objects in empty or absent states, we study the performance on prompts describing objects in full state (e.g., ``a tumbler full of water''). We provide these prompts in the suppl. material. We report the results in Table~\ref{tab:generalizability_results} column 1. Our results indicate that there is minimal difference in the performance of the models even after fine-tuning. This experiment highlights that our approach does not lead to catastrophic forgetting of objects in full states even though this is not explicitly represented in the finetuning data.


\subsection{Is synthetic data generation necessary?}
\input{sec/tables/different_data_sources_performance}
To evaluate whether synthetic data is essential for improving text-to-image generation, we extract 12K training examples from both the COCO~\cite{coco2014microsoft} dataset and a video dataset (VidOSC~\cite{vidosc}) that contain captions related to object states. For the real-world data, we filter the captions to retain only those whose associated prompts and images suggest a potential empty state, even if not explicitly described, as such samples are relatively scarce, as we discussed in the above sections. We then refine these captions, following the same process as our synthetic pipeline. The key difference between the synthetic dataset and the real-world dataset is that, for each filtered prompt in the real-world datasets, we directly use the corresponding image. In contrast, because the synthetic pipeline lacks corresponding real images and is based on a limited set of original prompts, we generate each prompt with multiple random seeds and perform reception on each generation.

Table~\ref{tab:data_source_comparison} compares the performance of models fine-tuned on these different data sources. Despite training all models on approximately 12K images for 400 steps, the synthetic dataset significantly outperforms the baseline. The modest gains from the real-world datasets suggest that, despite similar training set sizes, they contain relatively few high-quality samples that clearly represent object empty states. 
For a visual comparison of training samples across these datasets, please refer to the figure in the supplementary material.
In contrast, our synthetic dataset, specifically designed to capture varied object states, provides a much stronger fine-tuning signal, leading to a significant improvement in image-prompt alignment.
\input{sec/tables/fid_vs_clip_coco_performances}
\subsection{Performance impact on prompts not related to object states}
To verify whether the proposed framework impacts performance on prompts unrelated to object states, we leverage GenAI-Bench~\cite{li2024genai} and sample $50$ random prompts which are specifically outside the ``negation'' set. We test the performance of both the base and finetuned Stable Diffusion 2.1~\cite{Rombach_2022_CVPR} models when the synthetic data is generated using Stable Diffusion-1.5. Upon evaluating the generated images on such randomly sampled prompts, we observe that the base model has a GPT score of 10\% while our fine-tuned model has a performance of 16\%(+6 percentage points). This demonstrates that our approach improves a generative model's understanding of object states without deteriorating performance on other non-object state related prompts.


\subsection{Impact of recaptioning in the pipeline}
One of the modules in the synthetic data pipeline is the recaptioning segment, where we convert the template like prompts to passive voice prompts. For example, a prompt such as “An empty table.” can be recaptioned as “A table without any bottles on it” (see Fig.~\ref{fig:recaptioning_prompt}), shifting from a direct, template-like description to a more informal passive construction. We evaluate the importance of this step by finetuning Stable Diffusion 1.5~\cite{Rombach_2022_CVPR} and SDXL~\cite{podell2023sdxl,meng2021sdedit} with and without recaptioning the training data. We report the results in Table.~\ref{tab:recaptioning_ablation} and observe an improvement of $3$ and $1$ percentage points in GPT and VQA scores, respectively, in GenAI-Object-State. We also observe +8\% GPT score and +4\% VQA score points in Object-State-Bench. These results show that recaptioning to make prompts less-template like and more colloquial aligns them.
\input{sec/tables/recaptioning_ablation_performance}

\subsection{Evaluation on CommonsenseT2I dataset}
To verify if this pipeline instills commonsense in generating image from text, we have compared the change in performance on the CommonsenseT2I dataset~\cite{fu2024commonsense}. We test the performance of base and finetune models of Stable Diffusion Family~\cite{podell2023sdxl, Rombach_2022_CVPR, meng2021sdedit}, Flux~\cite{flux2024} and OmniGen~\cite{xiao2024omnigen} models on this dataset and we report the results in Table \ref{tab:commonsenset2i_perf}. Each prompt of the 150 pairs present in the dataset is considered as an independent prompt and so we get a total of 300 prompt description pairs. We use the prompt to generate image and we use the corresponding description to evaluate the generated images. Since these prompts do not exactly represent object states, we use a simpler prompt for GPT evaluation: ``You are an assistant that evaluates whether an image accurately represents a given prompt.  The provided caption is: ``prompt". Based on the caption, determine if the image correctly depicts the described content. Respond only with `yes' or `no' ". We see very little degradation in performance with an average of 3.5\% decrease in GPT score and 2.8\% decrease in VQA score. These results indicate that while this pipeline does not contribute towards model learning general commonsense, it does support the fact that this pipeline improves the model's understanding of object states in text while not destroying its knowledge in other domains.

\input{sec/tables/commonsenset2i_performance}

\subsection{User Study of Visual Quality}
\label{sec:appendix_user_study}
To verify that fine-tuning with our synthetic data does not degrade overall visual fidelity, we recruited 30 participants. Each participant was shown 50 paired samples which were generated using identical prompts and inference parameters from the original and the fine-tuned models. We asked the participants to choose which image looked better or `even” if indistinguishable.
Table~\ref{tab:user_quality} summarizes the win rates:
\begin{table}[ht]
  \centering
  \begin{tabular}{l r}
    \toprule
    Condition                   & Win Rate (\%) \\
    \midrule
    With tuning (ours)          & 52 \\
    Without tuning              & 46 \\
    Even                        & 2  \\
    \bottomrule
  \end{tabular}
  \caption{\textbf{Win rates comparison} from the image-quality user study.}
  \label{tab:user_quality}
\end{table}
As shown, 52\% of the time users preferred the fine‑tuned model, 46\% preferred the baseline, and only 2\% were ties.  This demonstrates that our synthetic‑data fine‑tuning does \textbf{not} introduce any undesirable artifacts or degrade visual quality, while  substantially improving generation correctness 

%% file: sec/figs/sd15_comparision.tex
\begin{figure}[ht!]
\begin{center}
    \includegraphics[width=1.0\linewidth]{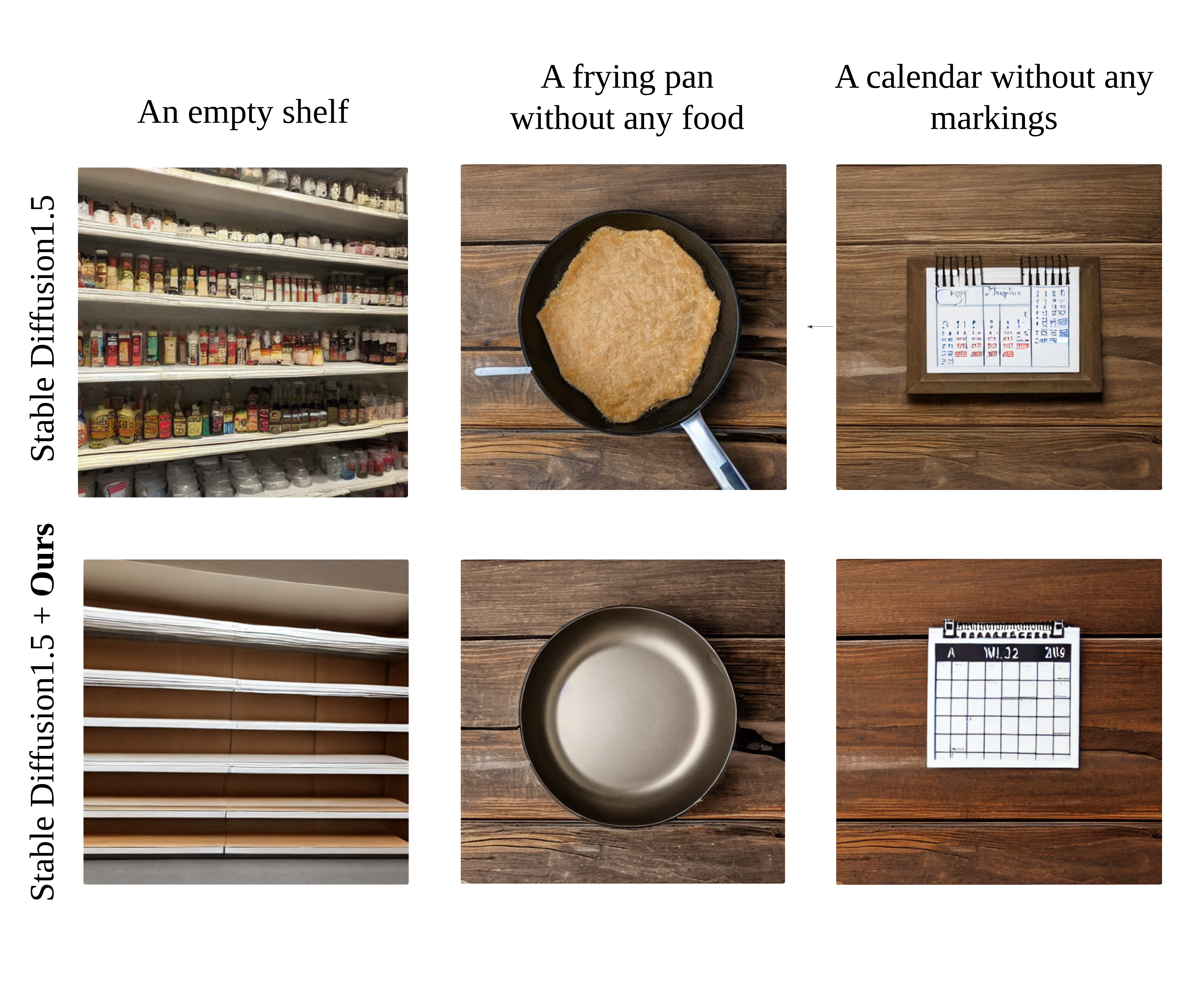}
\end{center}

\vspace{-6mm}
\caption{\textbf{Qualitative comparison of object state improvement for Stable Diffusion-1.5:} \textit{(top)} row shows the Stable Diffusion-1.5 baseline model, while the \textit{(bottom)} row displays fine-tuned with our synthetic data pipeline, yielding more precise object state representation.}
\label{fig:SD15_compare}
 \end{figure}

%% file: sec/figs/sd21_comparision.tex
\begin{figure}[ht!]
\begin{center}
\hspace*{-0.01\linewidth}
    \includegraphics[width=1.02\linewidth]{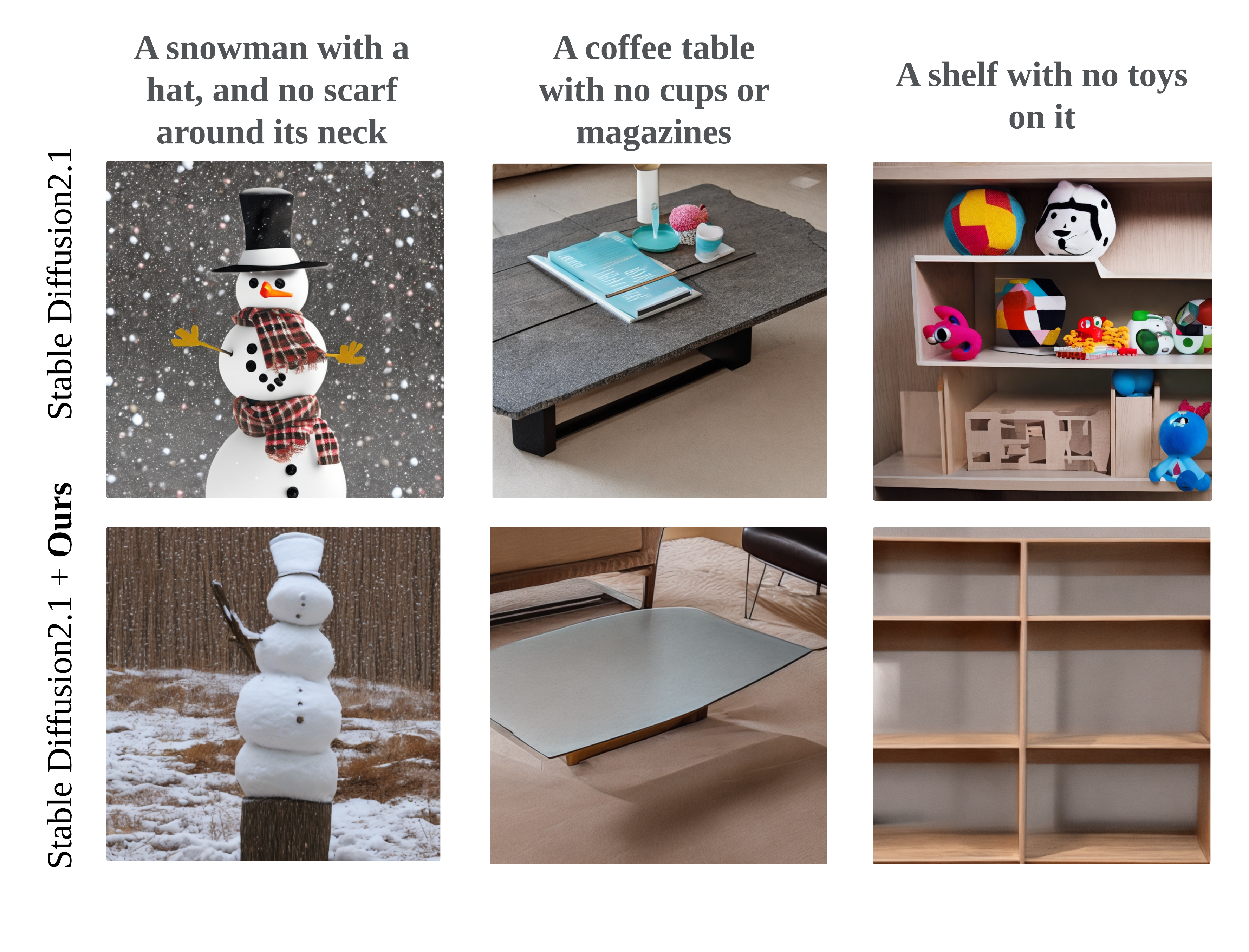}
\end{center}

\vspace{-6mm}
\caption{\textbf{Qualitative comparison of object state improvement for Stable Diffusion-2.1:} \textit{(top)} row shows the Stable Diffusion-2.1 baseline model, while the \textit{(bottom)} row displays fine-tuned with our synthetic data pipeline, yielding more precise object state representation.}

\label{fig:sd21_compare}
 \end{figure}

%% file: sec/figs/sdxl_comparision.tex
\begin{figure}[ht!]
\begin{center}
\hspace*{-0.06\linewidth}
    \includegraphics[width=1.04\linewidth]{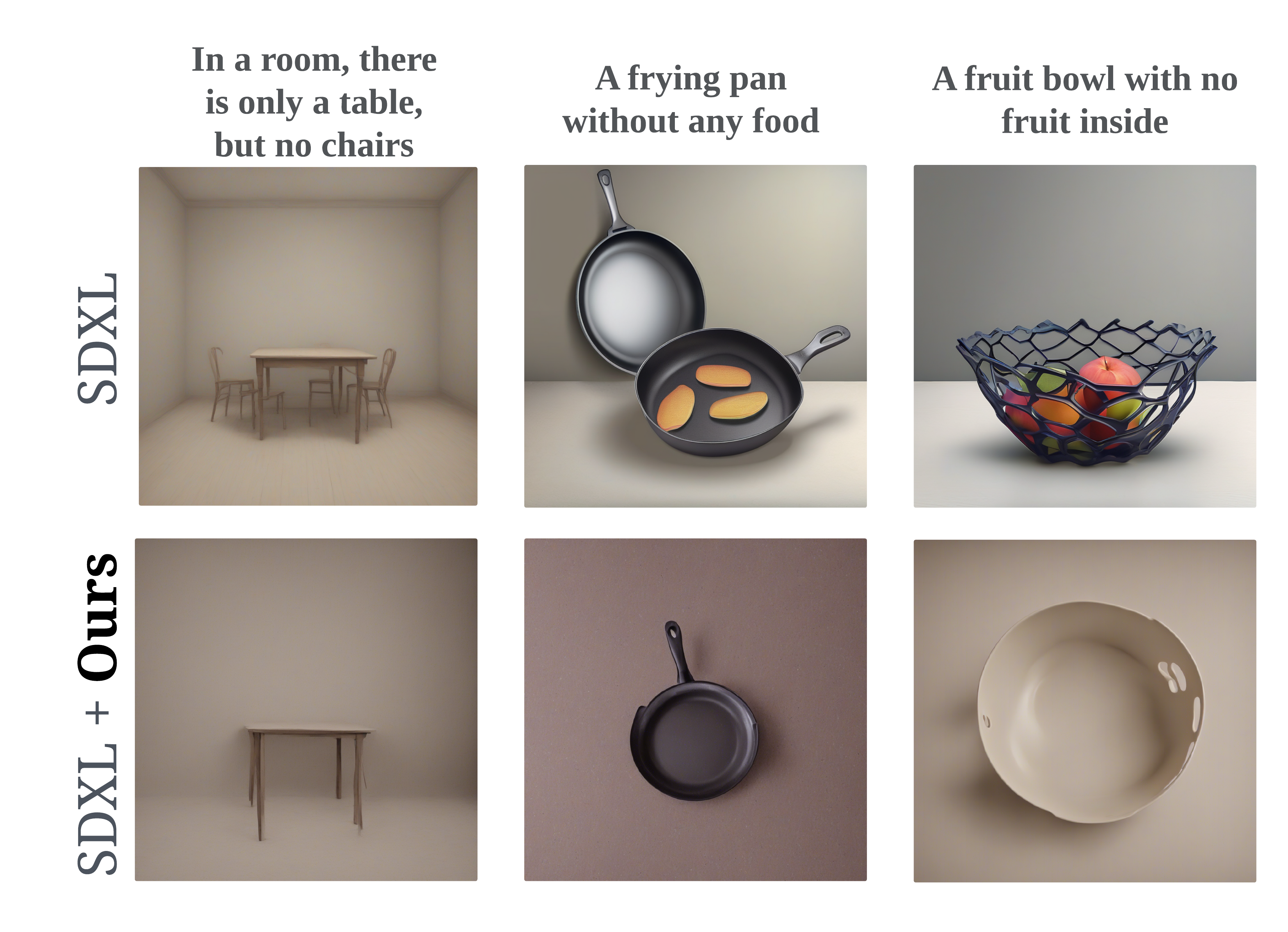}
\end{center}

\vspace{-6mm}

\caption{\textbf{Qualitative comparison of object state improvement for SDXL:} \textit{(top)} row shows the SDXL baseline model, while the \textit{(bottom)} row displays fine-tuned with our synthetic data pipeline, yielding more precise object state representation.}

\label{fig:sdxl_compare}
 \end{figure}

%% file: sec/figs/flux_comparision_fig.tex
\begin{figure}[ht!]
\begin{center}
    \includegraphics[width=0.97\linewidth]{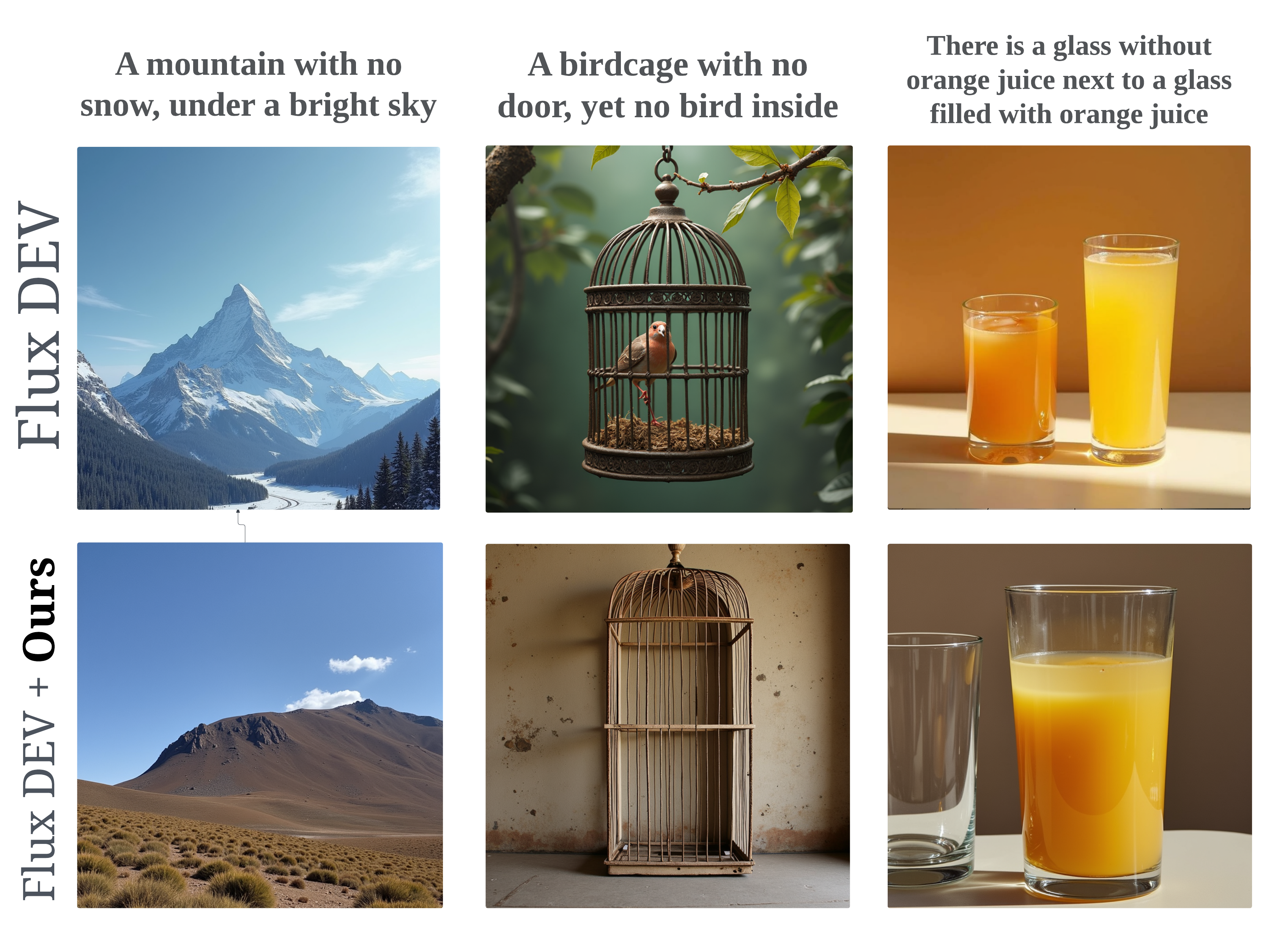}
\end{center}

\vspace{-6mm}
\caption{\textbf{Qualitative comparison of object state improvement for Flux Dev:} \textit{(top)} row shows the Flux Dev baseline model, while the \textit{(bottom)} row displays fine-tuned with our synthetic data pipeline, yielding more precise object state representation.}
\label{fig:flux_compare}
 \end{figure}

%% file: sec/figs/omnigen_comparision_fig.tex
\begin{figure}[ht!]
\begin{center}
    \includegraphics[width=0.97\linewidth]{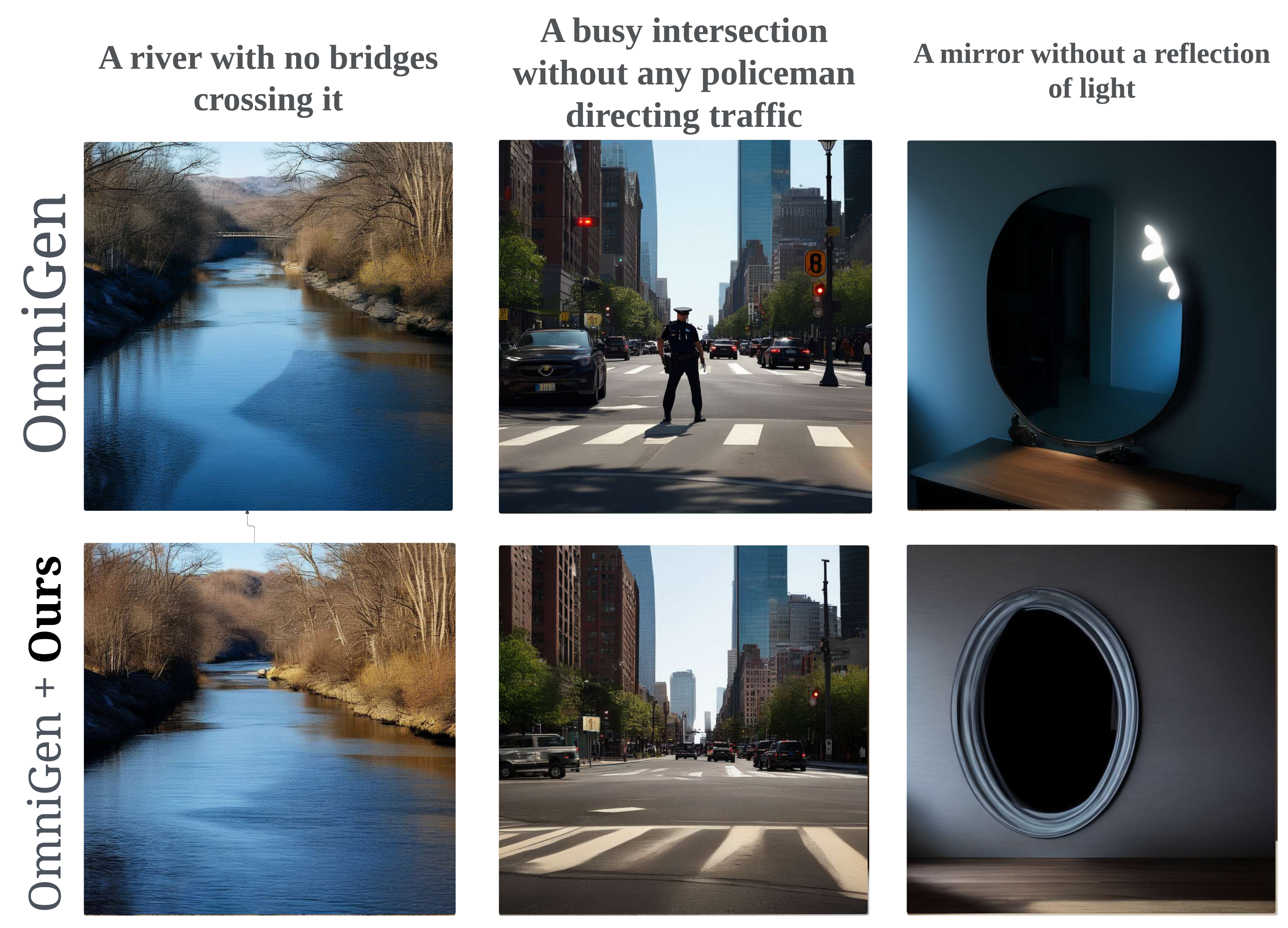}
\end{center}
\vspace{-2mm}
\caption{\textbf{Qualitative comparison of object state improvement for OmniGen:} \textit{(top)} row shows the OmniGen baseline model, while the \textit{(bottom)} row displays fine-tuned with our synthetic data pipeline, yielding more precise object state representation.}
\label{fig:omnigen_compare}
 \end{figure}

%% file: sec/4_1_implementation_details.tex
\subsection{Implementation Details}
\noindent \textbf{Synthetic data generation:} \label{sec:syn_pip_exp}
The synthetic data pipeline has multiple modules involved to ensure high quality training dataset and uses GPT-4o-mini~\cite{hurst2024gpt} in every step.
Specifically, we use 
GPT-4o--mini  
to generate about $3000$ different common objects and template-style prompts capturing these objects in empty or absent states. Next, we employ few-shot prompting technique~\cite{brown2020language} which has evidence to show better performance aligning with the prompted task in large language models.
For generating the synthetic images, we use Stable Diffusion 1.5~\cite{Rombach_2022_CVPR} 
for $30$ inference steps with a CFG scale of $5.0$. We choose a slightly lower CFG value than the default to ensure more diversity in the synthetic training data while adhering to the actual input prompt. We again use GPT 4o-mini~\cite{hurst2024gpt} to filter out images which incorrectly capture object states as mentioned in~\ref{sec:approach} and to rephrase the template-like prompts \ref{fig:recaptioning_prompt}. This process resulted in 7600 synthetic image-text pairs. 
\\

\input{sec/figs/recaptioning_prompt_fig}

\noindent \textbf{Evaluation benchmarks:} There exist very few public datasets that specifically focus on evaluating models on prompts capturing objects in varied physical states. The recently introduced GenAI-Bench~\cite{li2024genai}, and the subset of $347$ prompts belonging to the ``negation'' category is the closest to the scenario we study in this work. A sample prompt from this set is: ``the girl with glasses is drawing, and the girl without glasses is singing." Here the prompt challenges the model to generate both presence and absence of the same object (glasses) but on different people. However, the negation category also has prompts that test absence of attributes (instead of objects), e.g., ``a person with short hair is crying while a person with long hair is not.''
Thus, we manually go through these prompts and retain only those that are more aligned to our task, resulting in $214$ prompts. We call this subset as \textbf{GenAI-Object-State} dataset. 

Additionally, we manually curate a set of $200$ prompts, titled \textbf{Object-state-Bench}. This benchmark consists of two parts: one-half of the prompts are generated using the synthetic prompt generation pipeline and the other half is curated by human annotators tasked with describing common objects around them in empty or absent states. This design incorporates both machine-generated and human-authored descriptions, ensuring diversity and realistic linguistic variability for a robust evaluation of model performance.

\noindent \textbf{Evaluation metrics:} We quantify our model performance using the Visual Question Answering score (VQA-score) introduced by ~\citet{lin2024evaluating}. The metric utilizes a finetuned version of the Google's FLAN-T5-XXL model~\cite{FlanT5} with contrastive language-image pre-training~\cite{OpenAI_CLIP}. We use the default prompt given by the authors: ``Does this figure show ``prompt''? Please answer yes or no." Additionally, we also use OpenAI's GPT-4o-mini model~\cite{hurst2024gpt} for evaluation, where we query with an evaluation prompt specified in Fig.~\ref{fig:evaluation_prompt} and the generated image. The model returns a yes or no based on whether the object state has been correctly depicted.
\input{sec/figs/evaluation_prompt_fig}

%% file: sec/figs/recaptioning_prompt_fig.tex
\begin{figure}[t!]
    \centering
    \fbox{
        \begin{minipage}{0.95\linewidth}
            \textbf{Prompt for recaptioning into passive voive prompts:}
            \vspace{0.3cm}
            \textit{\\``The original prompt for the image is: `\{original\_prompt\}'. Please refine the prompt by specifying an absent object if it is not already mentioned, but avoid redundant descriptions of emptiness. Ensure the refined prompt naturally integrates the missing object without repeating words like `empty' or `vacant'. For example: `An empty table.' → `A table without any bottles on it.', 'A deserted park.' → `A park without any people.' If the original prompt is already sufficiently detailed, return it as is."}
        \end{minipage}
    }
   \caption{\textbf{System Recaptioning Prompt}: This figure shows the system prompt that transforms template-like prompts into passive voice. The examples instruct the model to enhance the prompt by adding a missing object and avoiding redundant emptiness descriptors.}
    \label{fig:recaptioning_prompt}
\end{figure}

%% file: sec/figs/evaluation_prompt_fig.tex
\begin{figure}[t!]
    \centering
    \fbox{
        \begin{minipage}{0.95\linewidth}
            \textbf{Prompt for evaluating generated images on representing absence or empty state of an object:}
            \vspace{0.3cm}
            \textit{\\``You are an assistant that evaluates whether an image correctly represents the `empty state' of an object as described in the caption. The caption is: \{original\_prompt\}. Specifically, check if the main object appears empty or unoccupied and confirm that the described absent object is not present in the image. Does the image accurately reflect both conditions? Return `yes' or `no'."}
        \end{minipage}
    }
    \caption{\textbf{System Prompt for Evaluation:} This figure presents the prompt used to assess whether a generated image accurately represents the absence or empty state of an object as described in the caption.}
    \label{fig:evaluation_prompt}
\end{figure}

%% file: sec/4_2_fine_tuning.tex
\subsection{Finetuning setup}
\noindent \textbf{Implementation Details}: 
We finetune Stable diffusion 1.5, 2.1~\cite{Rombach_2022_CVPR}, SDXL~\cite{podell2023sdxl,meng2021sdedit}, Flux.1 Dev~\cite{flux2024} and OmniGen~\cite{xiao2024omnigen} on the proposed framework. For the Stable Diffusion family of models, we use a guidance scale of $7.5$, which is their default value, for Flux DEV~\cite{flux2024}, we use a guidance scale of $3.5$ and for OmniGen, we use a guidance scale of $3$. Stable Diffusion 1.5 generates $512 \times 512$ dimensional output image while all other models generate $768 \times 768$ resolution. We infer Flux DEV version~\cite{flux2024} with $50$ inference timesteps 
as recommended in their documentation. For all other models, the number of inference time steps is $30$. 
We also ensure that the same seed of $1303$ (chosen arbitrarily) 
is used across all the prompts of the dataset for a given run for all the models during baseline testing. We finetune all models using Low Rank Adapters (LoRA)~\cite{hu2022lora}. The hyperparameters of LoRA are detailed in Appendix. 

\input{sec/tables/baseline_performances_table}

%% file: sec/tables/baseline_performances_table.tex
\begin{table}[t!]

\begin{center}
\resizebox{\linewidth}{!}{%
\begin{tabular}{lcc|cc}
\toprule
\multirow{2}{*}{\textbf{Method}} & \multicolumn{2}{c|}{\textbf{GenAI-Object-State}} & \multicolumn{2}{c}{\textbf{Object State Bench}} \\
\cmidrule(lr){2-3} \cmidrule(lr){4-5}
 & \textbf{GPT (↑)} & \textbf{VQA (↑)} & \textbf{GPT (↑)} & \textbf{VQA (↑)} \\
\midrule
Stable Diffusion 1.5 \cite{Rombach_2022_CVPR}              & 16\% & 45\% & 38\% & 42\% \\
Stable Diffusion 2.1 \cite{Rombach_2022_CVPR}   & 16\% & 47\% & 31\% & 40\% \\
SDXL \cite{podell2023sdxl,meng2021sdedit}                  & 15\% & 47\% & 33\% & 42\% \\
Flux DEV \cite{flux2024}                          & 11\% & 39\% & 19\% & 29\% \\
OminiGen \cite{xiao2024omnigen}              & 13\% & 40\% & 20\% & 31\% \\

\midrule
\rowcolor{cyan!10}Stable Diffusion 1.5 + Ours  & 21\% & 46\% & 54\% & 53\% \\
\rowcolor{cyan!10}Stable Diffusion 2.1 + Ours  & \textbf{23\%} & 49\% & 54\% & 55\% \\
\rowcolor{cyan!10}SDXL + Ours             & \textbf{23\%} & \textbf{52\%} & \textbf{56\%} & \textbf{58\%} \\
\rowcolor{cyan!10}Flux Dev + Ours         & 22\% & 50\% & \textbf{56\%} & 55\% \\ 
\rowcolor{cyan!10}OmniGen + Ours         &  \textbf{23\%} & 47\% & 44\% & 49\% \\ 
\bottomrule
\end{tabular}
}
\caption{\footnotesize{\textbf{Baseline vs finetuned results on different models}: GPT (↑) stands for GPT~\cite{hurst2024gpt} correct rate and VQA (↑) stands for VQA-Score~\cite{lin2024evaluating} (higher is better).} The cyan colored rows represent the performance of  fine-tuned models' with SD 1.5~\cite{Rombach_2022_CVPR} as synthetic data generator. The highlighted numbers indicate the highest performance for the corresponding dataset and metric.}
\label{tab:baseline_performances}
\vspace{-5mm}
\end{center}
\end{table}

%% file: sec/tables/finetune_performances.tex
\begin{table}[t]
\centering

\resizebox{\columnwidth}{!}{%
\begin{tabular}{c|c|c|c}
\toprule
\textbf{\shortstack{Tuned Generative\\Model}} & \textbf{\shortstack{Synthetic Data\\Generator}} & \textbf{\shortstack{GenAI-Object\\State}} & \textbf{\shortstack{Object State\\Bench}} \\
\midrule
\multirow{3}{*}{Stable Diffusion 1.5\cite{Rombach_2022_CVPR}} 
 & Stable Diffusion 1.5\cite{Rombach_2022_CVPR}  & 21\% & 53\% \\ 
 & SDXL\cite{podell2023sdxl,meng2021sdedit}               &  \textbf{24\%} &54\% \\
 & Stable Diffusion 2.1\cite{Rombach_2022_CVPR}  &  21\% &  56\% \\
\midrule
\multirow{3}{*}{SDXL\cite{podell2023sdxl,meng2021sdedit}} 
 & Stable Diffusion 1.5\cite{Rombach_2022_CVPR}   &  23\% &  53\% \\
 & SDXL\cite{podell2023sdxl,meng2021sdedit}       & 20\% & \textbf{58\%} \\
 & Stable Diffusion 2.1\cite{Rombach_2022_CVPR} &  22\% &  \textbf{58\%} \\
\midrule
\multirow{3}{*}{Stable Diffusion 2.1\cite{Rombach_2022_CVPR}} 
 & Stable Diffusion 1.5\cite{Rombach_2022_CVPR}  &  23\% &  56\% \\
 & SDXL\cite{podell2023sdxl,meng2021sdedit}             &  23\% &  55\% \\
 & Stable Diffusion 2.1\cite{Rombach_2022_CVPR}   &  22\% &  55\% \\
\bottomrule
\end{tabular}%
}
\caption{\textbf{Performance Comparison:} This table compares the performance of different synthetic data generators when used to fine-tune generative models on two benchmarks: GenAI-Object-State and Object State Bench. Results are reported as GPT evaluation scores (higher is better), with bold values indicating the best performance in each configuration.}
\label{tab:results_synthetic}
\vspace{-3mm}
\end{table}

%% file: sec/figs/partial_improvement_case.tex
\begin{figure}[ht!]
\vspace{-5mm}
\centering
\begin{subfigure}[b]{0.50\textwidth}
  \centering
  \includegraphics[width=\textwidth]{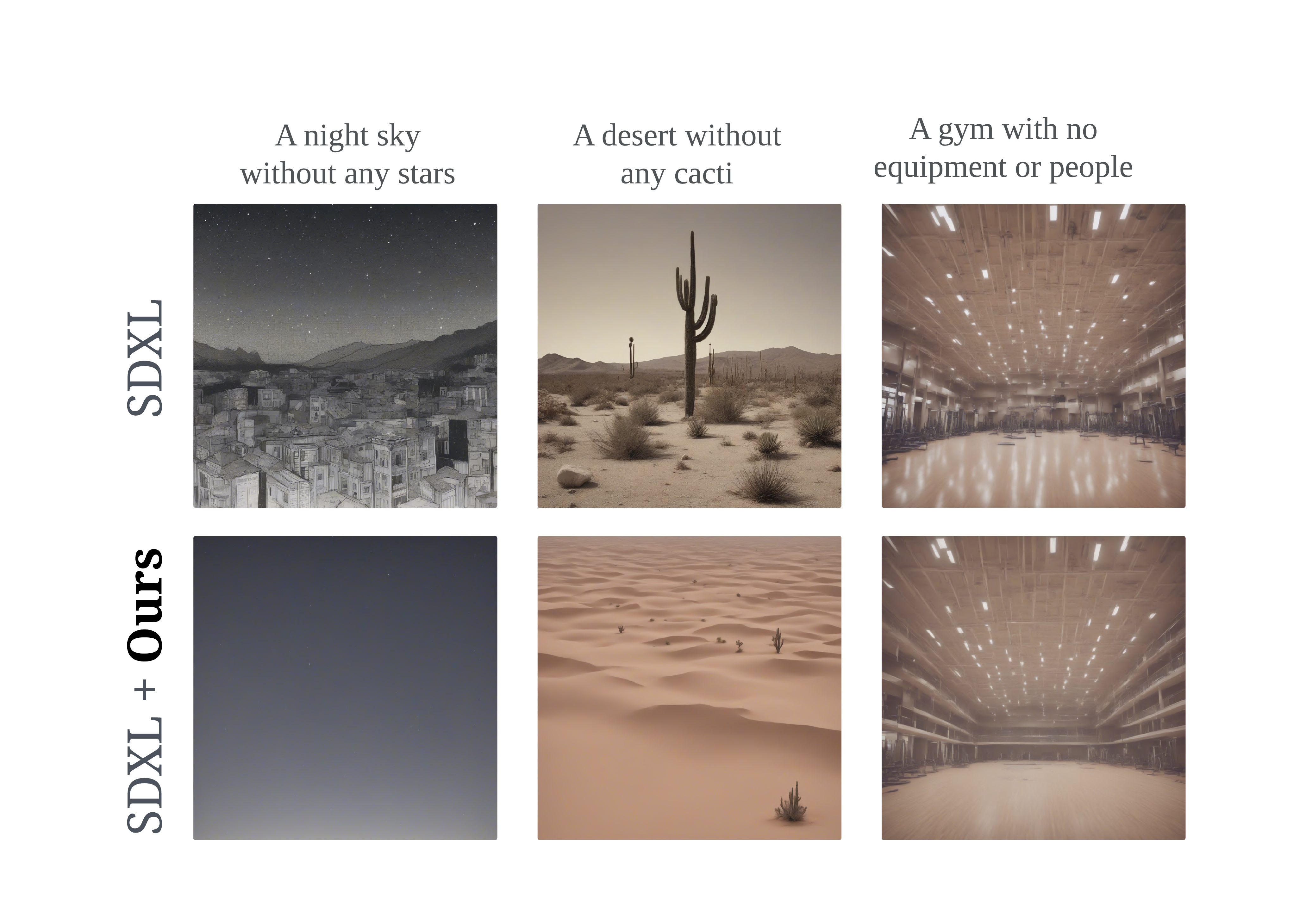}
  \caption{Tuned model shows improved object state, though not fully correct.}
  \label{fig:improvement_case}
\end{subfigure}
\vspace{-2mm}
\hfill
\hspace*{-0.05\linewidth}
\begin{subfigure}[b]{0.51\textwidth}
  \centering
  \includegraphics[width=\textwidth]{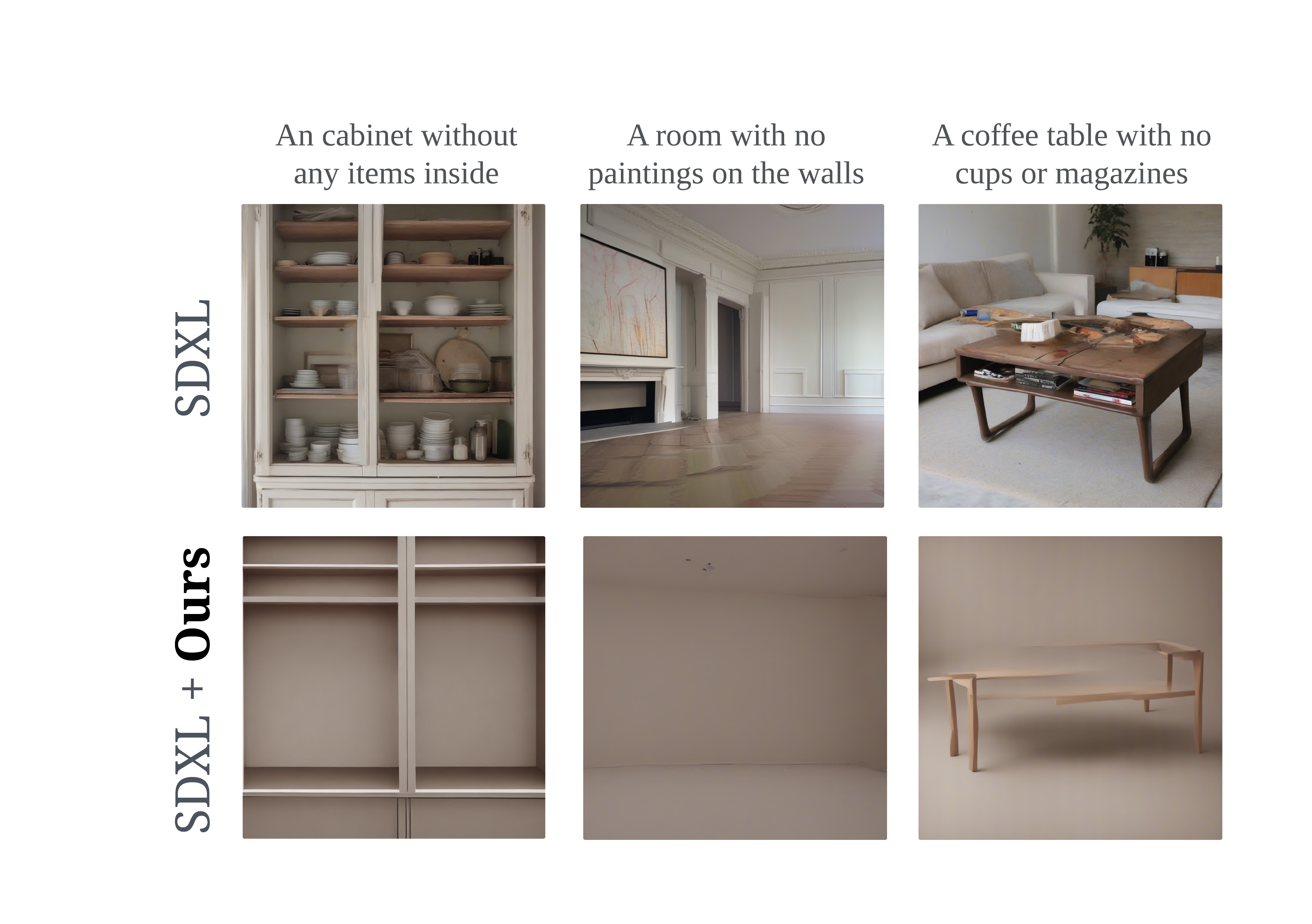}
  \caption{Tuned model overrepresents the empty state, shifting the object depiction.}
  \label{fig:overrepresentation_case}
\end{subfigure}
\caption{\textbf{Qualitative Tuning Effects:} Figure (a) shows cases where tuning improves the depiction of the object state (although with some imperfections), while Figure (b) illustrates instances where tuning overemphasizes emptiness, leading to a deviation from an accurate object representation.}
\label{fig:qualitative_tuning}
\vspace{-5mm}
\end{figure}

%% file: sec/tables/generalizability_performances.tex
\begin{table}[ht!]
\centering

\resizebox{\columnwidth}{!}{%
\begin{tabular}{c|cc|cc}
\toprule
\multirow{2}{*}{\textbf{Model}} & \multicolumn{2}{c|}{\textbf{\shortstack{Objects in\\ non-empty states}}} & \multicolumn{2}{c}{\textbf{\shortstack{Unseen \\ objects}}} \\
\cmidrule(lr){2-3} \cmidrule(lr){4-5}
 & \textbf{GPT (↑)} & \textbf{VQA (↑)} & \textbf{GPT (↑)} & \textbf{VQA (↑)} \\
\midrule
Stable Diffusion 1.5~\cite{Rombach_2022_CVPR}  & \textbf{40\%} & 68\% & 13\% & 44\% \\
Stable Diffusion 1.5~\cite{Rombach_2022_CVPR} + Ours & 39\% & \textbf{69\%} & \textbf{20\%} & \textbf{50\%} \\
\bottomrule
\end{tabular}%
}
\caption{\textbf{Evaluating Generalizability:} This table compares the performance of the baseline Stable Diffusion 1.5 with our fine-tuned variant on two evaluation sets: objects in non-empty states and unseen objects. GPT and VQA scores (higher is better) are reported for each category, with bold values indicating the best performance.}
\label{tab:generalizability_results}
\vspace{-5mm}
\end{table}

%% file: sec/tables/different_data_sources_performance.tex
\begin{table}[t!]

\centering
\resizebox{1.0\linewidth}{!}{%
\begin{tabular}{lccc}
\toprule
\textbf{Data Source} & \textbf{GPT (↑)} & \textbf{VQA (↑)}  \\
\midrule
Stable Diffusion 1.5\cite{Rombach_2022_CVPR} (Baseline)  & 38\% & 42\%  \\
COCO\cite{coco2014microsoft}        & 35\% & 42\%  \\
VidOSC \cite{vidosc}       & 36\% & 41\%  \\    
Stable Diffusion 1.5 + Synthetic Data (Ours)   & \textbf{54\%} & \textbf{53\%}  \\
\bottomrule
\end{tabular}
}
\caption{\textbf{Data Source Performance Comparison:} This table compares the performance of models fine-tuned on various data sources. Training on our synthetic dataset significantly boosts the GPT score (from 38\% to 54\%) and provides a modest improvement in VQA score over real-world datasets such as COCO~\cite{coco2014microsoft} and VidOSC~\cite{vidosc}.}
\label{tab:data_source_comparison}
\vspace{-3mm}
\end{table}

%% file: sec/tables/fid_vs_clip_coco_performances.tex
\begin{table}[ht!]
\centering
\label{tab:fid_clip}
\resizebox{\columnwidth}{!}{%
\begin{tabular}{lcc}
\toprule
\textbf{Generative Model} & \textbf{FID (↓)} & \textbf{CLIP-Score (↑)} \\
\midrule
Stable Diffusion 1.5~\cite{Rombach_2022_CVPR}   & 24.32  & 0.31 \\
Stable Diffusion 1.5~\cite{Rombach_2022_CVPR} + Ours  & 25.74  & 0.32 \\
\bottomrule
\end{tabular}
}
 \caption{\textbf{Effect of finetuning on synthetic data on visual quality:} Comparison of FID (lower is better) and CLIP-Score (higher is better) for Stable Diffusion 1.5 on COCO val2014 dataset, 10K randomly sampled subset.}
 \vspace{-3mm}
\end{table}

%% file: sec/tables/recaptioning_ablation_performance.tex
\begin{table}[ht!]
\centering
\resizebox{\columnwidth}{!}{%
\begin{tabular}{cc|cc|cc}
\toprule
\multirow{2}{*}{\textbf{Model}} & \multirow{2}{*}{\textbf{Recaptioning}} & \multicolumn{2}{c|}{\textbf{GenAI-Object-State}} & \multicolumn{2}{c}{\textbf{Object State Bench}} \\
\cmidrule(lr){3-4} \cmidrule(lr){5-6}
 &  & \textbf{GPT (↑)} & \textbf{VQA (↑)} & \textbf{GPT (↑)} & \textbf{VQA (↑)} \\
\midrule
\multirow{3}{*}{Stable Diffusion 1.5~\cite{Rombach_2022_CVPR}} 
 & N/A (baseline)  & 16\% & 45\% & 38\% & 42\% \\
 & No  & 17\% & 44\% & 42\% & 46\% \\
 & Yes  & 21\% & 46\% & 54\% & 53\%  \\
\midrule
\multirow{3}{*}{Stable Diffusion 2.1~\cite{podell2023sdxl,meng2021sdedit}} 
 & N/A (baseline)  & 16\% & 47\% & 31\% & 40\% \\
 & No  & 20\% & 48\% & 52\% & 52\% \\
 & Yes & 23\% & 49\% & 54\% & 55\% \\
\midrule
\multirow{3}{*}{SDXL~\cite{podell2023sdxl,meng2021sdedit}} 
 & N/A (baseline)  & 15\% & 47\% & 33\% & 42\% \\
 & No  & 20\% & 52\% & 45\% & 51\% \\
 & Yes  & 23\% & 52\% & 56\% & 58\% \\
\bottomrule
\end{tabular}%
}
\caption{\footnotesize{\textbf{Effect of Recaptioning on Model Performance:} This table presents a comparison of models fine-tuned with and without recaptioning. The results show that incorporating recaptioning improves both GPT and VQA scores on the GenAI-Object-State and Object State Bench, demonstrating its effectiveness in enhancing semantic alignment. Baseline scores are also included for reference.}}
\label{tab:recaptioning_ablation}
\end{table}

%% file: sec/tables/commonsenset2i_performance.tex
\begin{table}[ht!]
\centering
\resizebox{0.8\linewidth}{!}{%
\begin{tabular}{lccc}
\toprule
\textbf{Model} & \textbf{GPT (↑)} & \textbf{VQA (↑)}  \\
\midrule
Stable Diffusion 1.5~\cite{Rombach_2022_CVPR}  & 40\% & 59\%  \\
Stable Diffusion 2.1~\cite{Rombach_2022_CVPR}  & 41\% & 62\%  \\
SDXL~\cite{podell2023sdxl,meng2021sdedit}      & 44\% & \textbf{63\%}  \\
Flux DEV~\cite{flux2024}                       & 45\% & 62\%  \\
OmniGen~\cite{xiao2024omnigen}                 & 39\% & \textbf{63\%}  \\
\midrule
Stable Diffusion 1.5 + Ours                    & 39\% & 58\% \\
Stable Diffusion 2.1 + Ours                    & 36\% & 58\%  \\
SDXL + Ours                                    & \textbf{46\%} & 62\%  \\
Flux DEV + Ours                                & 40\% & 58\%  \\
OmniGen + Ours                                 & 35\% & 59\%  \\
\bottomrule
\end{tabular}
}
\caption{\textbf{Evaluation on CommonsenseT2I dataset:} This table compares the performance of the baseline models with our fine-tuned variants on CommonsenseT2I dataset~\cite{fu2024commonsense}. GPT and VQA scores (higher is better) are reported. Note that there is a slight decrease in performance after finetuning with the proposed pipeline.}
\label{tab:commonsenset2i_perf}
\vspace{-5mm}
\end{table}

%% file: sec/5_conclusion_future_work.tex
\section{Conclusion and Future Work}
To improve the physical object state representation in existing text to image generative systems, we propose a fully automatic pipeline to generate high-quality synthetic data and use it to finetune any text-to-image models. We demonstrate that our approach improves the holistic understanding of objects in diverse physical states via two evaluation metrics GPT score and VQA score~\cite{lin2024evaluating}. Future work entails exploring if such a data generation pipeline can be extended to other common failure models of image and video generative systems (compositional prompts, prompts involving generating text, generating objects of accurate counts, etc.) and exploring solutions with direct architectural tweaks to the model itself.

%% file: sec/appendices.tex
\clearpage  
\appendix   

\section*{Appendices}
\section{List of Unseen Objects}\label{appendix:object_list}
In this section, we provide the complete list of 100 novel objects used for evaluation. These objects were carefully selected and manually filtered to ensure they are distinct from the 3000 training objects. (The detailed list is provided here.)

\begin{figure}[htbp]
\centering
\small 
\begin{minipage}{\columnwidth}
\begin{multicols}{3}
\begin{itemize}[leftmargin=*, noitemsep, nolistsep]
\item birdhouse
\item meat locker
\item medicine bottle
\item print cartridge
\item shipping container
\item fuel can
\item bird's nest
\item fuel drum
\item shopping cart
\item squeeze bottle
\item photobook
\item paper towel holder
\item lipstick tube
\item sunglasses rack
\item jewelry roll 
\item wristwatch case
\item gadget case 
\item marionette theatre
\item drill holder
\item silicone mold
\item speech bubble
\item caviar tin
\item crisper drawer
\item resealable bag
\item hawser reel
\item luggage trunk
\item coffee tin
\item mason jar
\item picnic hamper
\item jewelry stand
\item picture frame
\item tackle pouch
\item flower vase
\item pool hall
\item exam room
\item motel room
\item kitchen closet
\item terracotta planter
\item patio chair
\item operating theater
\item animal cage
\item phone booth
\item quilt bag
\item dessert cup
\item sitting room
\item ring box
\item sleeping bag
\item flower vase with no water
\item traveling case
\item banana holder
\item rooftop garden bed
\item pressure cooker
\item guitar rack
\item sous vide container
\item cane basket
\item treehouse
\item planter
\item charcoal holder
\item dog treat tin
\item glaze bucket 
\item hair tie holder
\item bead organizer
\item scale pan
\item hammock frame
\item bulletin board
\item incense tray
\item book safe
\item picture rail
\item mantle clock case
\item snow globe
\item football bladder
\item card holder
\item protein container
\item chess board
\item animal trap
\item trailer
\item mannequin
\item candy mold
\item pet water fountain
\item vendor cart
\item film reel
\item stamp album
\item first-adi kit
\item toboggan
\item ice cream maker
\item hand towel rail
\item flower sac
\item toothbrush holder
\item control pannel
\item scout cabin
\item preserving jar
\item guitar case
\item sunhat box
\item noodle bowl
\item wristlet
\item salt grinder
\item cake pop stand
\item wine testing glass
\item gas lamp
\item billiard table
\end{itemize}
\end{multicols}
\end{minipage}
\caption{\textbf{List of 100 unseen objects} used for evaluation. These objects, which are distinct from the 3000 training objects, are detailed here for reproducibility.}
\label{fig:object_list}
\end{figure}

(Sec.\ref{sec:experiment}).

\section{List of Full state prompts}
In this section, we provide the full state prompts in Table ~\ref{fig:full_state_prompts} used for ablation study on performance on non-empty object states. These prompts were inspected manually to ensure there was no empty state reference.
\input{sec/figs/full_state_prompts}
\section{Comparison of Training Samples}
In this section, we provide a visual and qualitative comparison of training samples extracted from three different data sources: our synthetic dataset, COCO~\cite{coco2014microsoft}, and VidOSC~\cite{vidosc}. Our synthetic dataset is specifically generated to capture object empty states with clear and consistent imagery. In contrast, the COCO dataset often fails to clearly represent object absence: frequently, the described object is not the focal point, while the VidOSC dataset, derived from video frames, suffers from motion blur and inconsistent viewpoints. These limitations in real-world data help explain why models fine-tuned on our synthetic dataset perform significantly better in generating accurate object states.

\begin{figure}[t!]
\centering
\includegraphics[width=\linewidth]{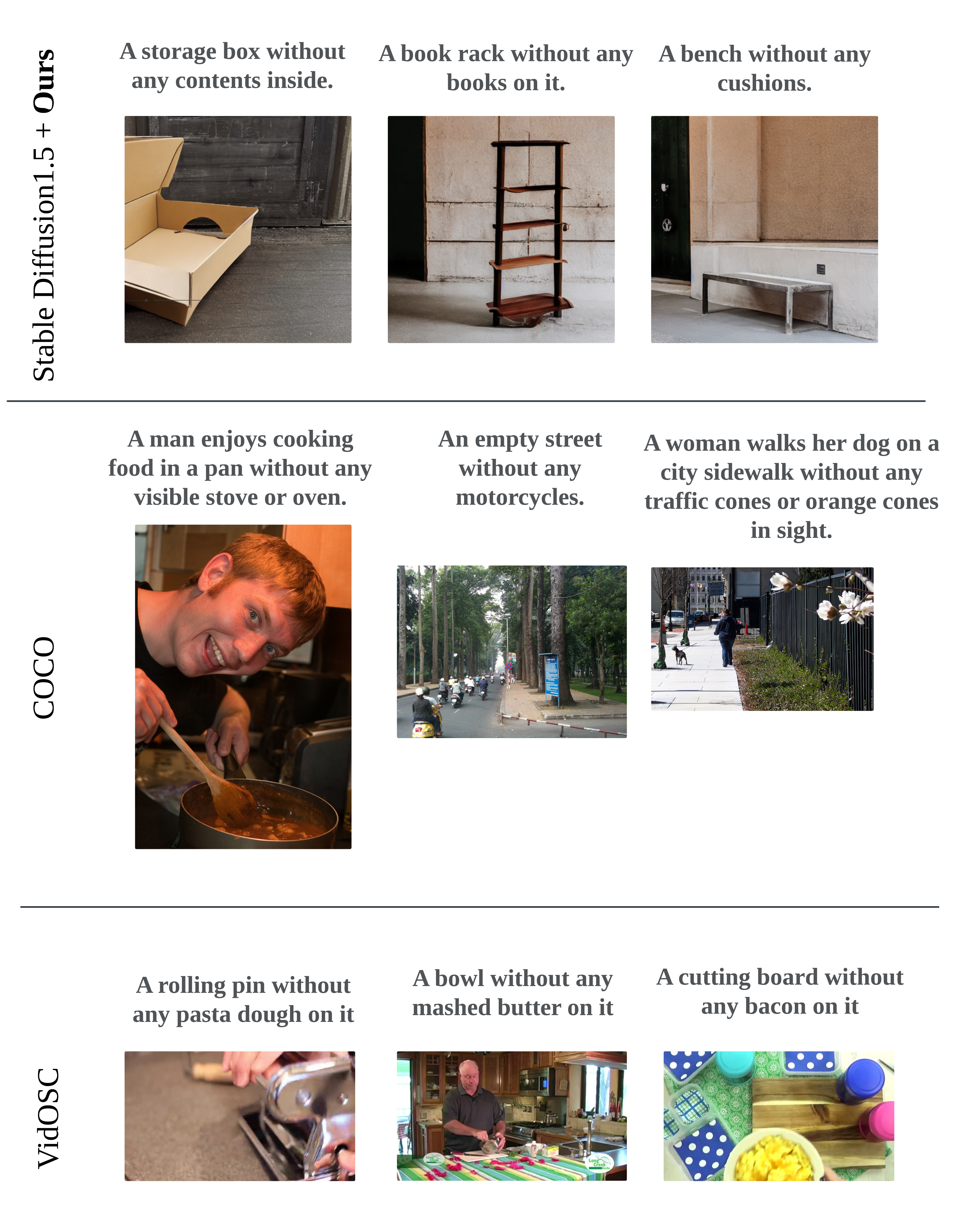}
\caption{\textbf{Training Sample Comparison from Different Data Sources:} The top row shows samples from our synthetic dataset, the middle row displays samples from COCO~\cite{coco2014microsoft}, and the bottom row presents samples from VidOSC~\cite{vidosc}. Our synthetic dataset clearly captures object empty states with focused and consistent imagery. In contrast, COCO samples often fail to clearly depict the absence of an object, with the described object not being the focal point, while VidOSC samples suffer from motion blur and inconsistent viewpoints. These factors contribute to the superior performance of our synthetic dataset.}
\label{fig:sample_comparison}
\end{figure}

\section{Hyperparameters for LoRA in finetuning}
We outline the hyperparameters used for LoRA during fine-tuning of the family of Stable Diffusion models~\cite{Rombach_2022_CVPR}, Flux.1 DEV~\cite{flux2024} and OmniGen~\cite{xiao2024omnigen} in table \ref{tab:hyperparameters_lora}.

\input{sec/tables/lora_hyperparams}

\section{Additional Qualitative Examples of Object State Failures}

In this appendix, we provide further qualitative results that illustrate additional failure cases of closed-source text-to-image generation models when handling object state prompts. These examples reinforce our observations and emphasize the need for improved model training and data recaptioning.
\input{sec/figs/advanced_model_failure_more}

\section{Impact of Tuning Steps on Accuracy}

In this section, we analyze how varying the number of fine tuning steps affects accuracy on the GenAI‑Object‑State benchmark. Table~\ref{tab:tuning_steps} summarizes GPT scores for each model at 200, 400, and 800 tuning steps. For Stable Diffusion 1.5, the score rises from 20\% at 200 steps to 21\% at 400 steps, then falls to 18\% at 800 steps. Stable Diffusion 2.1 improves from 18\% to 23\% between 200 and 400 steps, before a slight drop to 22\% at 800. SDXL shows a steadier increase, moving from 23\% at 200 steps to 24\% at 800. These trends suggest that around 400 tuning steps offer the best balance between semantic alignment and stability. Based on validation on a 50‑sample set, we adopt 400 steps for all subsequent experiments.

\begin{table}[t!]
\centering
\resizebox{\linewidth}{!}{%
\begin{tabular}{lccc}
\toprule
\textbf{Model}                    & \textbf{200 Steps} & \textbf{400 Steps} & \textbf{800 Steps} \\
\midrule
Stable Diffusion 1.5              & 20\%               & 21\%               & 18\%               \\
Stable Diffusion 2.1              & 18\%               & 23\%               & 22\%               \\
SDXL                              & 23\%               & 23\%               & 24\%               \\
\bottomrule
\end{tabular}%
}
\caption{GPT evaluation scores (in percentage) for different tuning steps on the GenAI-Object-State benchmark.}
\label{tab:tuning_steps}
\end{table}



%% file: sec/figs/full_state_prompts.tex
\begin{figure*}[htbp]
\centering
\small 
\begin{multicols}{3}
\begin{enumerate}[leftmargin=*, noitemsep, nolistsep]
\item A full bottle of water is placed on the table.
\item The cup is filled to the brim with hot coffee.
\item The plate is loaded with a delicious meal.
\item The bowl is full of fresh fruit.
\item The glass is filled with orange juice.
\item The jar is packed with homemade jam.
\item The container is filled with rice.
\item The box is stuffed with chocolates.
\item The bag is filled with groceries.
\item The wallet is thick with cash.
\item The suitcase is packed with clothes for the trip.
\item The backpack is filled with school supplies.
\item The envelope is stuffed with important documents.
\item The fuel tank is completely full, ready for a long drive.
\item The trash can is overflowing with garbage.
\item The sink is full of dirty dishes.
\item The bathtub is filled with warm, soapy water.
\item The fridge is stocked with fresh food.
\item The freezer is packed with frozen meals.
\item The oven is full of baking cookies.
\item The pan is filled with sizzling vegetables.
\item The pot is bubbling with hot soup.
\item The dish rack is full of clean plates.
\item The storage box is packed with winter clothes.
\item The wardrobe is filled with dresses and suits.
\item The bookshelf is packed with novels and textbooks.
\item The laundry basket is full of dirty clothes.
\item The washing machine is loaded with clothes.
\item The dryer is tumbling a full load of laundry.
\item The pencil case is filled with pens and markers.
\item The toolbox is stocked with hammers and screwdrivers.
\item The drawer is stuffed with office supplies.
\item The file cabinet is filled with paperwork.
\item The purse is heavy with personal items.
\item The shopping cart is loaded with groceries.
\item The refrigerator drawer is filled with fresh vegetables.
\item The spice rack is stocked with herbs and spices.
\item The medicine cabinet is filled with bottles of pills.
\item The candy jar is brimming with sweets.
\item The flower vase is full of fresh roses.
\item The aquarium is teeming with colorful fish.
\item The tea kettle is filled with boiling water.
\item The thermos is full of hot coffee.
\item The lunchbox is packed with sandwiches and snacks.
\item The picnic basket is overflowing with food and drinks.
\item The trash bag is full and needs to be taken out.
\item The egg carton is completely full.
\item The gas cylinder is filled with propane.
\item The rain barrel is full after the storm.
\item The bathtub is overflowing with bubbles.
\item The hard drive is full of stored files.
\item The email inbox is filled with unread messages.
\item The car trunk is packed with luggage.
\item The bread basket is full of warm rolls.
\item The coffee pot is filled with fresh-brewed coffee.
\item The pet food bowl is full for dinner time.
\item The ice cube tray is full and ready to freeze.
\item The cup holder is filled with soda cans.
\item The suitcase pocket is stuffed with travel essentials.
\item The fishing net is full of fresh catch.
\item The raincoat pockets are filled with small items.
\item The coin purse is full of loose change.
\item The fruit basket is overflowing with apples and bananas.
\item The measuring cup is filled with flour.
\item The battery pack is fully charged.
\item The balloon is filled with helium.
\item The notepad is full of handwritten notes.
\item The chalkboard is covered with writing.
\item The gift bag is stuffed with presents.
\item The music playlist is full of favorite songs.
\item The wine cellar is stocked with vintage bottles.
\item The parking lot is completely full.
\item The stadium is packed with cheering fans.
\item The toy chest is overflowing with stuffed animals.
\item The makeup bag is full of beauty products.
\item The tool shed is stocked with gardening equipment.
\item The bakery display case is filled with fresh pastries.
\item The cookie jar is full of chocolate chip cookies.
\item The seed packet is full of flower seeds.
\item The pet carrier is filled with cozy blankets.
\item The luggage rack is stacked with heavy suitcases.
\item The fishing bucket is full of water and fish.
\item The scrapbook is filled with memories.
\item The classroom board is covered with notes.
\item The violin case is packed with accessories.
\item The music stand is filled with sheet music.
\item The bike basket is loaded with fresh groceries.
\item The file folder is stuffed with reports.
\item The bread bin is stocked with fresh loaves.
\item The lemonade pitcher is full and ready to serve.
\item The attic is packed with old furniture and boxes.
\item The beach bag is full of towels and sunscreen.
\item The hospital bed is occupied with a patient.
\item The rain boot is filled with water after the storm.
\item The marshmallow jar is overflowing with sweets.
\item The milk carton is completely full.
\item The Christmas stocking is filled with gifts.
\item The dog’s food bowl is filled with kibble.
\item The holiday suitcase is packed with vacation clothes.
\item The bus is completely full of passengers.

\end{enumerate}
\end{multicols}
\caption{List of 100 full state prompts used for ablation study. }
\label{fig:full_state_prompts}
\end{figure*}

%% file: sec/tables/lora_hyperparams.tex
\begin{table}[ht]
\centering
\caption{Hyperparameters used for fine-tuning with LoRA.}
\resizebox{\columnwidth}{!}{%
\begin{tabular}{lccc}
\toprule
\textbf{Hyperparameter} & \textbf{SD family~\cite{Rombach_2022_CVPR}} & \textbf{Flux.1 DEV~\cite{flux2024}} & \textbf{OmniGen~\cite{xiao2024omnigen}}\\
\midrule
LoRA Rank & 4 & 16 & 16\\
Resolution & 512 & 512 & 512\\
Center Crop & True & True & False\\
Random Flip & True & False & False\\
Mixed Precision & fp16 & bf16 & bf16\\
Allow TF32 & True & False & -\\
Training Batch Size & 32 & 8 & 8\\
Gradient Accumulation Steps & 1 & 1 & 1\\
Gradient Checkpointing & True & True & -\\
Learning Rate & 1e-04 & 1e-04 & 1e-04\\
Max Gradient Norm & 1 & 1 & -\\
\bottomrule
\end{tabular}%
}
\label{tab:hyperparameters_lora}
\end{table}

%% file: sec/figs/advanced_model_failure_more.tex
\begin{figure*}[ht!]
\centering
\includegraphics[scale=0.13]{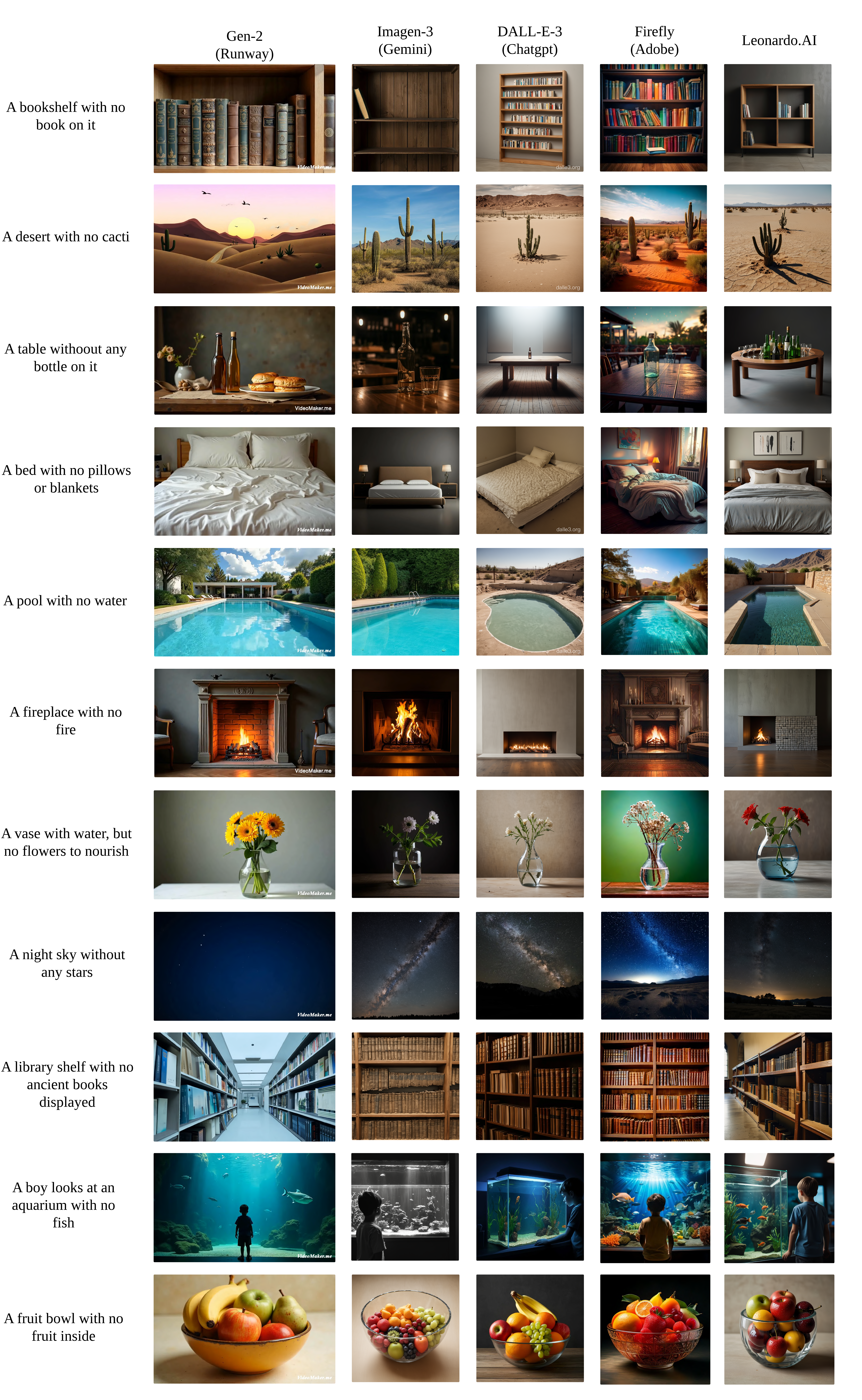}
\caption{Additional qualitative examples of object state failures in advanced text-to-image generation models. For example, the figure shows a bed without pillows or blankets, a fireplace with no fire, a pool with no water, a vase filled with water but missing flowers, a night sky devoid of stars, a library shelf lacking ancient books, an aquarium without fish, and a fruit bowl with no fruit. These deficiencies are observed in outputs from models such as Runway-Gen-3, Imagen-3 (Gemini), DALL-E-3 (ChatGPT), Firefly (Adobe), and Leonardo.AI.}
\label{fig:advanced_fail2}
\end{figure*}